\documentclass{ecai} 
\usepackage{latexsym}
\usepackage{amsthm,amsmath,amssymb}
\usepackage{booktabs}
\usepackage{enumitem}
\usepackage{color}

\usepackage{url}
\usepackage[hidelinks]{hyperref}
\usepackage{graphicx,subfig}
\usepackage{algorithm}
\usepackage{algorithmic}
\usepackage{makecell}
\usepackage{multirow}
\usepackage{array}
\usepackage{pifont}
\usepackage{subfig}
\usepackage[dvipsnames, svgnames, x11names]{xcolor}

\newcommand{\BibTeX}{B\kern-.05em{\sc i\kern-.025em b}\kern-.08em\TeX}

\definecolor{red2}{RGB}{240,191,211}
\definecolor{deepred}{RGB}{219,95,146}
\definecolor{blue2}{RGB}{205,226,247}
\definecolor{deepblue}{RGB}{88,161,230}
\begin{document}

%%%%%%%%%%%%%%%%%%%%%%%%%%%%%%%%%%%%%%%%%%%%%%%%%%%%%%%%%%%%%%%%%%%%%%%%

\begin{frontmatter}

%%% Use this command to specify your submission number.
%%% In doubleblind mode, it will be printed on the first page.

\paperid{154} 

%%% Use this command to specify the title of your paper.

\title{Adapt PointFormer: 3D Point Cloud Analysis via Adapting 2D Visual Transformers}

%%% Use this combinations of commands to specify all authors of your 
%%% paper. Use \fnms{} and \snm{} to indicate everyone's first names 
%%% and surname. This will help the publisher with indexing the 
%%% proceedings. Please use a reasonable approximation in case your 
%%% name does not neatly split into "first names" and "surname".
%%% Specifying your ORCID digital identifier is optional. 
%%% Use the \thanks{} command to indicate one or more corresponding 
%%% authors and their email address(es). If so desired, you can specify
%%% author contributions using the \footnote{} command.

\author[A,B]{\fnms{Mengke}~\snm{Li}}
\author[A,B]{\fnms{Da}~\snm{Li}}
\author[A,B]{\fnms{Guoqing}~\snm{Yang}}
\author[C]{\fnms{Yiu-ming}~\snm{Cheung}}
\author[B]{\fnms{Hui}~\snm{Huang}\thanks{Corresponding author.}}

\address[A]{ Guangdong Laboratory of Artificial Intelligence and Digital Economy (SZ), Shenzhen, China}
\address[B]{VCC, College of Computer Science and Software Engineering, Shenzhen University, Shenzhen, China}
\address[C]{Department of Computer Science, Hong Kong Baptist University, Hong Kong SAR, China}

%%% Use this environment to include an abstract of your paper.

% \maketitle
\begin{abstract}
Pre-trained large-scale models have exhibited remarkable efficacy in computer vision, particularly for 2D image analysis.
However, when it comes to 3D point clouds, the constrained accessibility of data, in contrast to the vast repositories of images, poses a challenge for the development of 3D pre-trained models.
%Furthermore, the process of retraining large models is labor-intensive and time-consuming.
This paper therefore attempts to directly leverage pre-trained models with 2D prior knowledge to accomplish the tasks for 3D point cloud analysis.
Accordingly, we propose the Adaptive PointFormer (APF), which fine-tunes pre-trained 2D models with only a modest number of parameters to directly process point clouds, obviating the need for mapping to images.
Specifically, we convert raw point clouds into point embeddings for aligning dimensions with image tokens.
Given the inherent disorder in point clouds, in contrast to the structured nature of images, we then sequence the point embeddings to optimize the utilization of 2D attention priors.
To calibrate attention across 3D and 2D domains and reduce computational overhead, a trainable PointFormer with a limited number of parameters is subsequently concatenated to a frozen pre-trained image model. 
Extensive experiments on various benchmarks demonstrate the effectiveness of the proposed APF.
% The source code is temporarily available in the supplementary material.
The source code and more details are available at 
\href{https://vcc.tech/research/2024/PointFormer}{\textcolor{blue}{\textit{https://vcc.tech/research/2024/PointFormer}}}.
\end{abstract}

%Code link: https://github.com/944104439/APF

\end{frontmatter}

\section{Introduction}
\label{sec:intro}
Compared with the traditional paradigm of training neural networks from scratch~\cite{li2024graph,pang2024heterogeneous}, pre-trained self-attention-based models~\cite{vaswani2017attention}, represented by BERT~\cite{devlin2018bert} and visual transformer (ViT)~\cite{Dosovitskiy21vit}, have shown significant improvement in natural language processing (NLP), image recognition, and related domains.
Transformer-based architecture has also been introduced for point cloud analysis in several studies~\cite{zhao2021point,guoMH2021pct,choe2022pointmixer} and has shown remarkable progress. 
Subsequently, the novel parameter-efficient fine-tuning paradigm (PEFT)~\cite{yu2023visual}, has been introduced to harness the rich prior knowledge and powerful representational capabilities inherent in pre-trained models for a wide array of downstream tasks. 
Recently, multiple 3D pre-trained models have been developed, such as OcCo~\cite{wangHC2021occo}, point-BERT~\cite{yu2022point}, and point-MAE~\cite{pangYT22PointMAE}, to name a few.
However, despite these advancements, a significant scarcity persists in the availability of extensive pre-trained models tailored for 3D point cloud analysis.
This scarcity is attributed to the considerably higher costs and labor-intensive efforts associated with the acquisition of accurately labeled 3D data, in contrast to the relative abundance of labeled data available in the domains of images and language. 
For example, OcCo is pre-trained on ModelNet40~\cite{wuZR20153d}, a dataset comprising 12,311 synthesized CAD objects from 40 categories.
In the realm of images, for instance, there are numerous well-trained transformer-based models, such as ViT-Base~\cite{Dosovitskiy21vit}, comprising 86 million parameters trained on a dataset of 14 million images, and CLIP~\cite{radford2021clip}, trained with 400 million image and text pairs. 
Given this scenario, a question arises: \textit{Can we directly leverage 2D prior knowledge for the analysis of 3D point clouds?} 
If feasible, the wealth of inexpensive and readily accessible 2D data, coupled with pre-trained models, holds the potential to substantially enhance the methods for point cloud analysis.

\begin{figure}[!tb]
\subfloat[ModelNet40]{ 
    \includegraphics[width=0.5\linewidth]{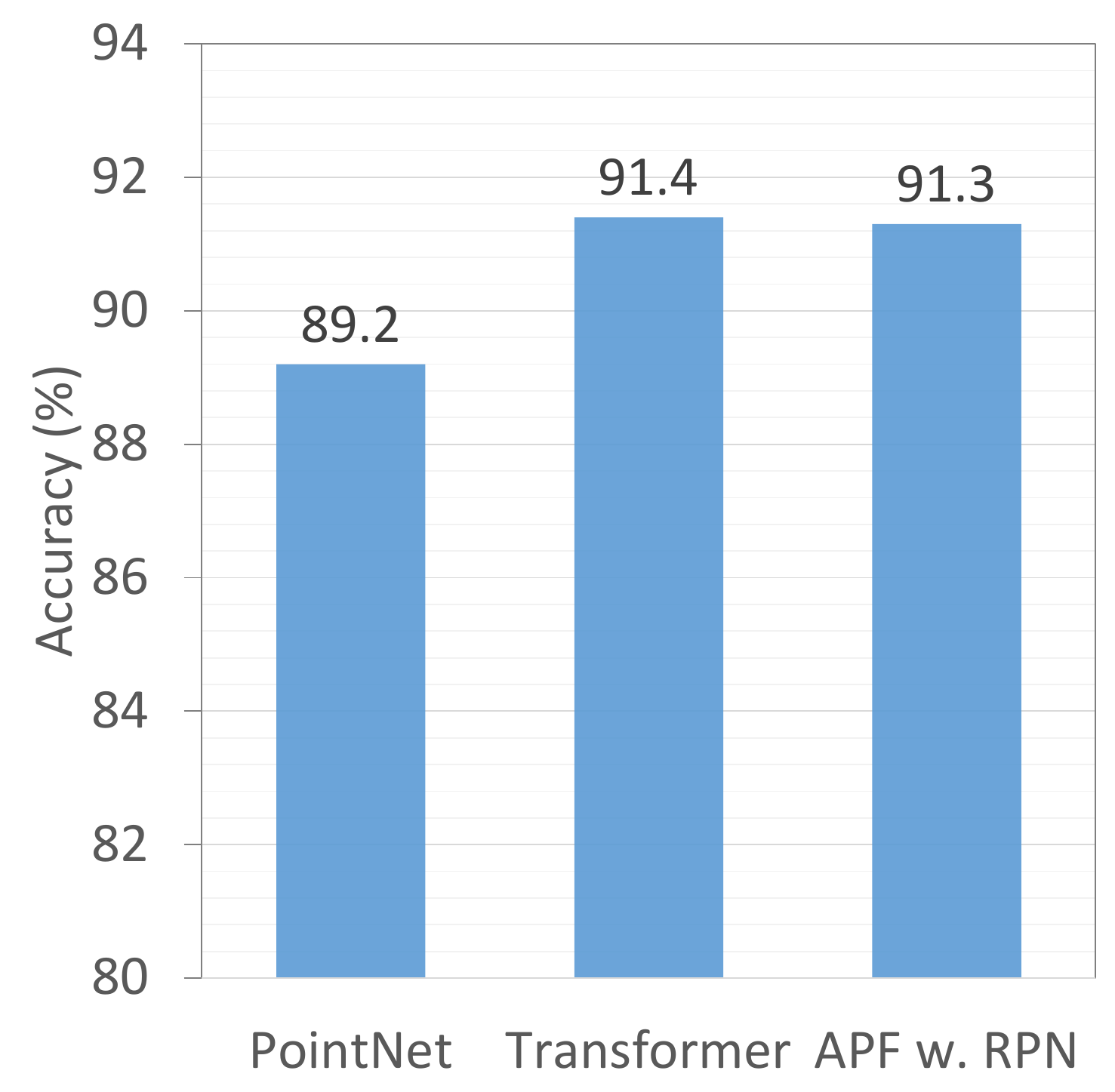}
    \label{fig:intro_MN40}
    }
\subfloat[ScanObjectNN]{
    \includegraphics[width=0.5\linewidth]{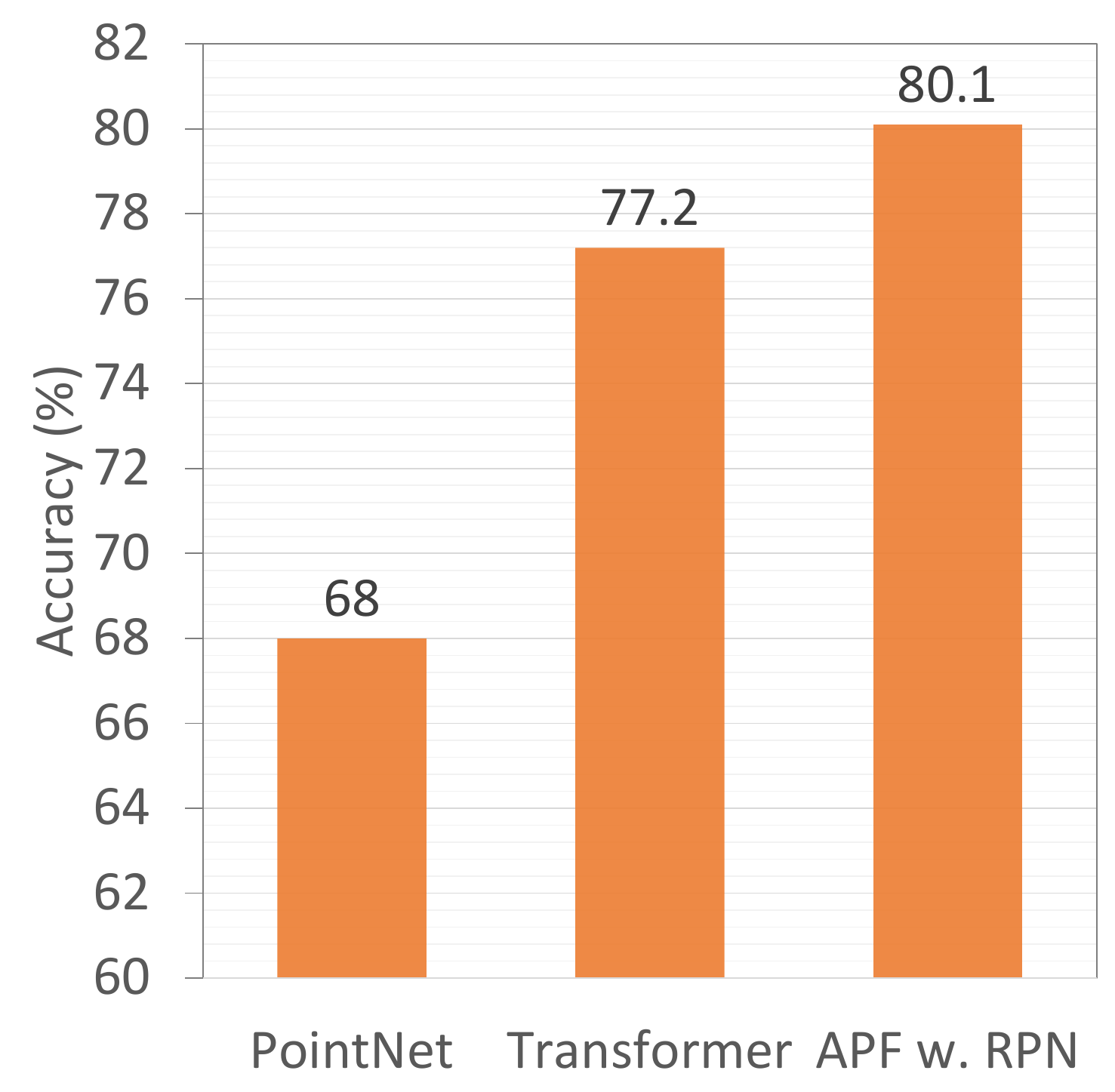}
    \label{fig:intro_SONN}
    }    
\vspace{6pt}
\caption{Performance comparison. APF w. RPN denotes our proposed APF architecture employing random lightweight PointNet.}
\vspace{18pt}
\label{fig:intro_acc_comp}
\end{figure}

The affirmative response is encapsulated in the work of Wang et al.~\cite{Ziyi21P2P}, who proposed Point-to-Pixel Prompting (P2P). 
This approach stands as the first attempt to transfer pre-trained knowledge from the 2D domain to the 3D domain.
Nevertheless, P2P requires mapping the point cloud into images, a process characterized by pathological mapping that inevitably leads to the loss of inherent information within point clouds.
Joint-MAE~\cite{GuoZQLH23JointMAE} investigates the geometric correlation between 2D and 3D representations. 
Employing a joint encoder and decoder architecture with modal-shared and model-specific decoders facilitates cross-modal interaction. 
Essentially, both P2P and joint-MAE leverage the point cloud and its corresponding projection to explore image knowledge.

We first devise an empirical study to further investigate the viability of directly applying image priors in point cloud analysis.
A randomly initialized lightweight PointNet (RPN) is utilized for aligning the dimensions of point clouds with that of image tokens and then obtaining the random point embedding.
The dimension alignment network is fixed during training.
Subsequently, a pre-trained 2D transformer undergoes fine-tuning with the sequenced random point embedding as input.
%Subsequently, a pre-trained 2D transformer undergoes stdfine-tuning technique based on AdaptFormer~\cite{chen2022adaptformer}, referred to as PointFormer, utilizing the RPE as input.
The results are shown in Figure~\ref{fig:intro_acc_comp}.
It can be observed that, compared to the model trained on 3D data from scratch, the fine-tuned 2D model attains higher accuracy.
Therefore, the attention derived from pre-trained models on 2D images exhibits efficacy in analyzing 3D point clouds.

To this end, this paper proposes a novel approach named Adapt PointFormer (APF), which utilizes pre-trained image models for the direct processing of point clouds, thereby adapting 2D image prior knowledge to 3D point clouds.%avoiding the ``one-to-many" 3D-2D projection.
To further effectively leverage 2D self-attention, APF renders the dimension alignment network to be trainable and incorporates point embedding sequencing. 
To better calibrate the attention of point clouds and image priors, a fine-tuning technique based on AdaptFormer~\cite{chen2022adaptformer}, referred to as PointFormer is introduced.
Extensive experiments are conducted on various downstream tasks to demonstrate the effectiveness of APF.

In summary, our main contributions are:
\begin{itemize}
    \item We investigate the potential of the pre-trained image model in 3D point cloud analysis and reveal that directly leveraging 2D priors with minimal fine-tuning can outperform models trained on 3D data from scratch.
    \item We propose APF, a framework that fine-tunes 2D pre-trained models for direct application to 3D point cloud analysis. 
    It consists of a point embedding module and a point sequencer for feature alignment, followed by a PointFormer module with a minimal number of trainable parameters for attention calibration.  
    \item We conduct extensive experiments on diverse 3D downstream tasks, which demonstrates the superior performance of APF compared to existing methods.
\end{itemize} 
\section{Related work}
\label{sec:related_work}
%In this paper, we focus primarily on weakly-supervised semantic segmentation of urban point clouds and review some related works that include supervised point cloud processing, semantic segmentation of the urban point cloud, and weakly-supervised semantic segmentation methods.

%-------------------------------------------------------------------------
\subsection{3D Point Cloud Analysis}

\noindent\textbf{CNN-based Methods.}
Since the introduction of PointNet~\cite{qi2017pointnet}, there has been a flourishing development of deep learning-based approaches in the realm of point cloud processing over the past few years.
These methods can be categorized into three groups based on the representations of point clouds: voxel-based~\cite{Liu2019pointvoxel,Shi2020CVPR}, projection-based~\cite{ranH2022surface,li2023bevdepth}, and point-based~\cite{guo2020deep,qianG2022pointnext}.  %projection-based: suHM2015multi,goyalAL2021revisiting
Voxel-based methods entail the voxelization of input points into regular voxels, utilizing CNNs for subsequent processing. 
However, these methods tend to incur substantial memory consumption and slower runtime, particularly when a finer-grained representation is required~\cite{guo2020deep}.
Projection-based methods encompass the initial conversion of a point cloud into a dense 2D grid, treated thereafter as a regular image, facilitating the application of classical methods to address the problems of point cloud analysis. 
However, these methods heavily rely on projection and back-projection processes, presenting challenges, particularly in urban scenes with diverse scales in different directions.
In contrast, point-based methods, directly applied to 3D point clouds, are the most widely adopted. 
Such methods commonly employ shared multi-layer perceptrons or incorporate sophisticated convolution operators~\cite{qi2017pointnet,QiNIPS2017pointnet2,Wang2019Dynamic,ThomasH2019KPConv}. 
In recent years, hybrid methods such as PVCNN~\cite{Liu2019pointvoxel} and PV-RCNN~\cite{Shi2020CVPR}, which combine the strengths of diverse techniques, have achieved notable advancements.
%-------------------------------------------------------------------------

\noindent\textbf{Self-Attention-based Methods.}
Self-attention operations~\cite{vaswani2017attention} have been adopted for point cloud processing in several studies~\cite{zhao2021point,guoMH2021pct,choe2022pointmixer}. 
For example, the point transformer~\cite{zhao2021point} and point cloud transformer (PCT)~\cite{guoMH2021pct} have introduced self-attention networks~\cite{vaswani2017attention} to improve the capture of local context within the point clouds.
Afterward, a plethora of methods based on the self-attention architecture have been proposed.
PointMixer~\cite{choe2022pointmixer} enhances self-attention layers through inter-set and hierarchical-set mixing.
TokenFusion~\cite{wang2022multimodal} initially fuses tokens from heterogeneous modalities with point clouds and images, subsequently forwarding the fused tokens to a shared transformer, allowing the learning of correlations among multimodal features.
AShapeFormer~\cite{li2023ashapeformer} utilizes multi-head attention to effectively encode information pertaining to object shapes. This encoding capability can be seamlessly integrated with established 3D object detection methodologies.
Exploiting pre-trained transformer models is also a promising way.
P2P~\cite{Ziyi21P2P} employs a lightweight DGCNN~\cite{Wang2019Dynamic} for the conversion of point clouds into visually rich and informative images, which serves to facilitate the utilization of pre-trained 2D knowledge.
Point-BERT~\cite{yu2022point} constructs point cloud tokens that represent various geometric patterns, resembling word tokens. 
Subsequently, pre-trained language models, represented by BERT~\cite{devlin2018bert}, can be applied to downstream tasks such as object classification, part segmentation, and related applications.

\subsection{Parameter-Efficient Fine-Tuning}
Recent advancements for 2D visual recognition incorporate the pre-trained Transformer models, exemplified by models like CLIP~\cite{radford2021clip} and ViT~\cite{Dosovitskiy21vit}.
Parameter-efficient fine-tuning (PEFT)~\cite{yu2023visual}, similar to model reprogramming or adversarial reprogramming~\cite{ElsayedGS19Adversarial} in the field of adversarial learning is designed to capitalize on the representational capabilities inherited from pre-trained models. 
PEFT strategically fine-tunes only a few parameters to achieve better performance in various downstream tasks that are different from the pre-trained models~\cite{chen2023understanding}.
Representative methods, including prompt tuning (PT)~\cite{LesterAC21PT}, adapters~\cite{houlsby2019parameter}, and Low-rank adapter (LoRA)~\cite{HuSWALWWC22LoRA} were initially designed for the purpose of incorporating language instructions into the input text for language models.
LoRA~\cite{HuSWALWWC22LoRA} applies parameter tuning within the multi-head self-attention module in transformers.
He \textit{et al.}~\cite{HeZMBN22ad} introduced adapters into the field of computer vision.
Bahng \textit{et al.}~\cite{bahng2022exploring} first defined the ``visual prompt", mirroring the ``prompt" in NLP.
Subsequently, PEFT in the realm of computer vision, such as visual prompt tuning (VPT)~\cite{jia2022visual}, visual adapters~\cite{chen2022adaptformer,nie2023pro}, and visual LoRA~\cite{shi2023parameter}, to name a few, exhibit outstanding performance with minimal training parameters, reduced epochs, and substantial performance enhancements.
Recently, there have been proposed numerous PEFT methods for fine-tuning 3D pre-trained models~\cite{{zha2023instance,tang2024point,zhou2024dynamic}}.
Any2Point~\cite{liu2024robomamba} proposes to utilize a PEFT method for fine-tuning pre-trained models with any modality, leveraging prior spatial knowledge.
This approach demonstrates the efficacy of pre-trained knowledge of different modalities in enhancing 3D understanding.
\section{Methodology}
\subsection{Motivation}
Self-attention-based architecture, commonly referred to as transformer, has demonstrated noteworthy advancements in the analysis of point clouds~\cite{guoMH2021pct,zhao2021point}.
Nonetheless, the transformer architecture~\cite{vaswani2017attention,devlin2018bert,Dosovitskiy21vit} exhibits inferior performance in comparison to CNNs when trained from scratch on a midsize dataset like imageNet-1K~\cite{Yuan21Tokens2Token}.
In contrast to the more readily accessible 2D image data, the acquisition and annotation of 3D point cloud data imposes considerable financial and temporal burdens~\cite{longFY23pointclustering} due to its irregular and inhomogeneous character.
This disparity poses a challenge in training transformer-based networks for point cloud analysis.
We, therefore, propose to utilize PEFT technology, which demands less training data.
However, existing pre-training models predominantly rely on 2D image data. 
To effectively utilize these pre-trained models, the calibration of dimensions and attention for 3D point clouds with 2D image prior becomes crucial.

\begin{figure*}[t]
    % \centering
    \begin{minipage}[t]{0.76\linewidth}
        \centering
        \includegraphics[width=\linewidth]{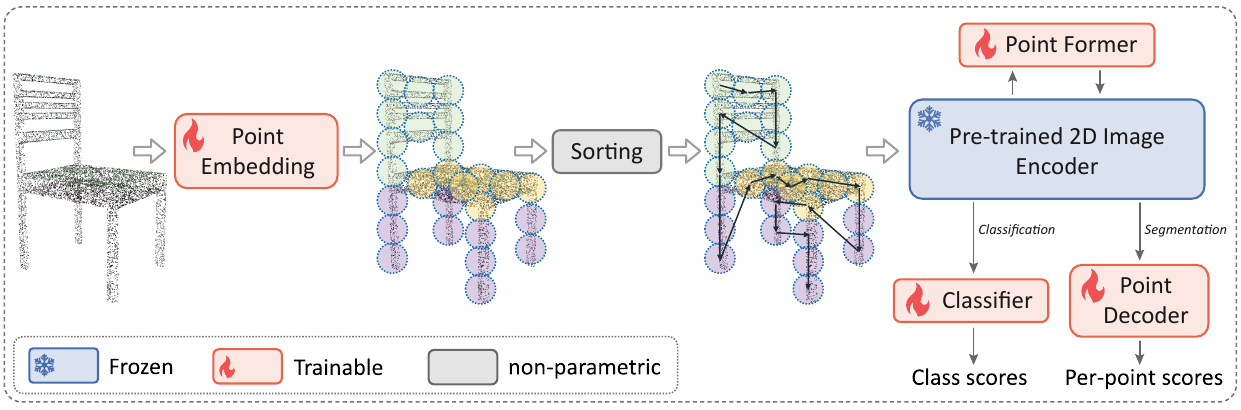}
        \caption{The pipeline of our proposed APF.}
        \label{fig:pipeline}
    \end{minipage}
    \hfill
    \begin{minipage}[t]{0.21\linewidth}
        \centering
        \includegraphics[width=\linewidth]{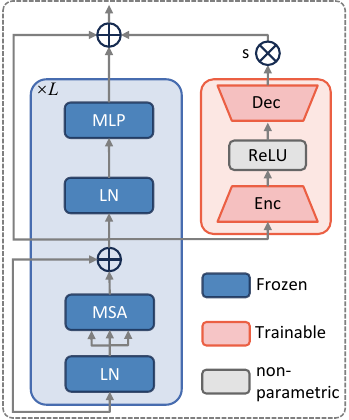}
        \caption{The framework of PointFormer.}
        \vspace{0pt}
        \label{fig:PointFormer}
    \end{minipage}
    \vspace{6pt}
\end{figure*}

\subsection{Preliminaries}
\noindent\textbf{2D Visual Transformer.} 
A visual transformer (ViT) model comprises an embedding layer and multiple transformer blocks.
For an input image $x^I$, the model first partitions $x^I$ into $m$ patches, forming a set $\{x^I_i\}_{i=1}^m$.
These patches are then embedded into sequences of $d^I$-dimensional vectors, denoted as $E^I_0=\texttt{Embed} \left( \left[x^I_1, x^I_2, \cdots, x^I_m \right] \right)$, where $E^I_0\in \mathbb{R}^{m \times d}$. 
$E^I_0$ is subsequently fed into $L$ blocks $\{\phi^{(l)}\}_{i=1}^L$ within the transformer model.
We use superscript $(l) $ to denote the index of the block.
Formally, this procedural description can be mathematically expressed as:
\begin{align}
        z_i^{I,(0)} &= \texttt{Embed}\left(x^I_i \right)+e_i, \label{eq:img_emd1}\\
        \left[z_\text{cls}^{I,(l)}, \mathcal{Z}^{I,(l)} \right] &= 
        \phi^{(l)}\left( \left[z_\text{cls}^{I,(l-1)}, \mathcal{Z}^{I,(l-1)} \right] \right)   \label{eq:img_emd2}  
\end{align}  
where $z_i^{I,(0)}\in \mathcal{R}^d$ and $e_i \in \mathcal{R}^d$ denote the image path embedding and positional embedding, respectively.
$\mathcal{Z}^{I,(l)} = [ z_1^{I, (l)}, z_2^{I, (l)}, \cdots, z_m^{I, (l)}]$.
$z_\text{cls}^{I,(l)}$ is an additional learnable token for classification. 
$\phi^{(l)}$ is composed of multi-head self-attention ($\texttt{MSA}$) and a MLP layer ($\texttt{MLP}$) with layer normalization ($\texttt{LN}$)~\cite{ba2016layer} and residual connection~\cite{he2016deep}. 
Specifically, $\phi^{(l)}$ is composed by:
\begin{equation} \label{eq:block}
    \begin{cases}
    \tilde{z}_i^{I,(l)} =  \texttt{MSA}^l\left( z_i^{I,(l-1)} \right) + z_i^{I,(l-1)}\\  
    z_i^{I,(l)} = \texttt{MLP}^l \left( \texttt{LN}\left(\tilde{z}_i^{I,(l)} \right) \right) + \tilde{z}_i^{I,(l)}
    \end{cases}.
\end{equation}
A singular self-attention within $\texttt{MSA}^l$ is calculated by the softmax-weighted interactions among the input query, key, and value tokens obtained by three different learnable linear projection weights. 
%A singular self-attention within $\texttt{MSA}^l$ is calculated by:
%\begin{equation}
%    \texttt{Att} \left( Q^{(l)}, K^{(l)}, V^{(l)} \right) = \texttt{SoftMax} \left( \dfrac{ Q^{(l)} {K^{(l)}}^T} {\sqrt{d}} \right) V^{(l)},
%\end{equation}
%where $Q^{(l)}, K^{(l)}$ and  $V^{(l)}$ are the input query, key, and value tokens obtained by three different learnable linear projection weights denoted by $\mathbf{W}_Q$, $\mathbf{W}_K$ and $\mathbf{W}_V$, respectively.
Finally, the class prediction is achieved by a linear classification head.

\noindent\textbf{Raw Point Grouping.}
Given an input point cloud $\mathcal{P} \in \mathbb{R}^{N \times {(d'+C)}}$, where $N$ represents the number of unordered points, denoted as $\mathcal{P}=\left[x^P_1, x^P_2, \cdots, x^P_N\right]$ and $x^P_i \in \mathbb{R}^{d'+C}$ with $d'$-dim coordinates and $C$-dim point feature, we first employ iterative farthest point sampling (FPS) to sample a subset of points $\mathcal{P}_s = \left[x^P_1, x^P_2, \cdots, x^P_{N_s} \right] \in \mathbb{R}^{N_s \times (d'+C)}$.
Subsequently, the $k$-nearest neighbors $\mathcal{P}_g = \left[ \left\{x^P_{1,j} \right\}_{j=1}^k,\left\{x^P_{2,j} \right\}_{j=1}^k, \cdots, \left\{x^P_{N_s,j} \right\}_{j=1}^k \right] \in \mathbb{R}^{N_s \times k \times (d'+C)}$ for each point are identified, wherein each group $\left\{x^P_{i,j} \right\}_{j=1}^k$ within $\mathcal{P}_g$ corresponds to a local region around the centroid point $x^P_{i}$, and $k$ represents the number of points adjacent to the $N_s$ centroid points.
Following this, embedding $\mathcal{P}_g$ becomes necessary to leverage the pre-trained 2D ViT structure.

\subsection{Point Embedding \& Sequencing}
\label{sec:pnt_emb}
Point embedding converts the grouped raw points into a structured and representative embedding for enhancing the utilization and calibration with the 2D image tokens.
We implement a lightweight network ($\texttt{Point\_Embed}$) to obtain the point embedding:
\begin{equation}
\label{eq:pnt_emb}
    z^{P,(0)}_i = \texttt{Point\_Embed}\left( \mathcal{X}_i^{P} \right),
\end{equation}
where $\texttt{Point\_Embed}$ can take various forms such as pointNet~\cite{qi2017pointnet}, pointNeXt~\cite{qianG2022pointnext} and pointMLP~\cite{maXQ22PointMLP}, to name a few.
The input point $x_i^{P}$ is from $\mathcal{P}_g$.
We use $\mathcal{X}_i^{P}$ to represent the set of $k$ neighboring points $\left\{x^P_{i,j} \right\}_{j=1}^k$ around $x^P_{i}$ for simplicity.
To seamlessly integrate with the 2D pre-trained ViT, the dimension of point embedding should align with image embedding in Eq.~(\ref{eq:img_emd1}).
Specifically, $ z^{P,(0)}_i \in \mathbb{R}^d$.
Eventually, the embedding representation of an input point cloud $\mathcal{P}$ for feeding into pre-trained 2D ViT is $\mathcal{Z}^{P,(0)} = \left[ z^{P,(0)}_1, z^{P,(0)}_2, \cdots, z^{P,(0)}_{N_s} \right] $.

\begin{figure}[!tb]
    \centering
    \includegraphics[width=0.7\linewidth, height = 0.35\linewidth]{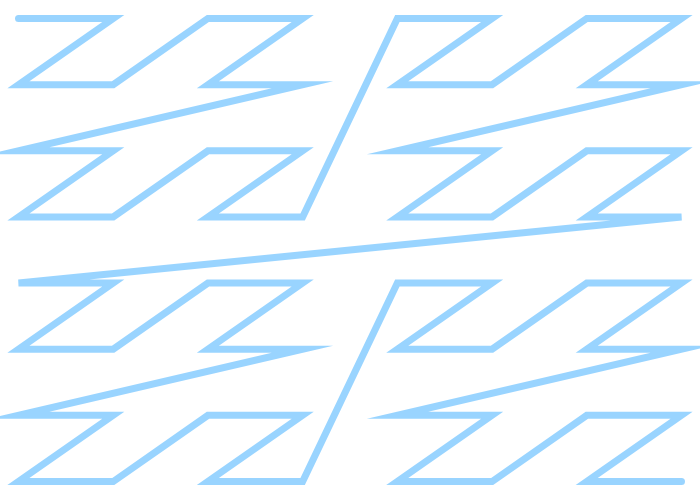}
    \vspace{6pt}
    \caption{Schematic of Morton-order curve.}
    \label{fig:z_order}
    \vspace{20pt}
\end{figure}

The inherent unordered nature is one of the most significant properties of point clouds~\cite{qi2017pointnet}, making it different from pixel arrays in image data.
Merely aligning the dimension of embeddings is insufficient to fully leverage the attention-related priors of a 2D pre-trained model.
We introduce a 3D token sequencer that leverages Morton-order~\cite{morton1966computer} to sequence the point embedding:
\begin{equation}
    \mathcal{O} = \texttt{Morton\_Order}\left( \mathcal{P}_s \right),
\end{equation}
where $\mathcal{O} \in \mathbb{R}^{N_s \times 1}$ is the order of input point sets.
$\texttt{Morton\_Order}$ is achieved by: 
1) Representing the coordinates of a point in binary. 
2) Interleaving the bits of these binary numbers. 
3) Convert the interleaved binary number back to a decimal value, referred to as Morton value (or Z value).
The schematic of the Morton-order curve is shown in Figure~\ref{fig:z_order}.
We sequence the point embedding obtained by Eq~(\ref{eq:pnt_emb})  according to Morton-order:
\begin{equation}
    \mathcal{Z}^{P}_s = \mathcal{Z}^{P}\left[ \mathcal{O} \right].
\end{equation}
For simplicity, we omit the superscript indicating which block the input belongs to.
Subsequently, the transformer-based model is utilized to acquire point tokens.

\subsection{PointFormer}
The transformer-based architecture is more data-hungry than CNN-based ones~\cite{Yuan21Tokens2Token}.
%In comparison to image data, the quantity of three-dimensional data is relatively limited, which causes overfitting and cannot fully release the ability of the transformer-based model.
In comparison to image data, the availability of 3D data is relatively constrained, resulting in issues such as overfitting and a limited realization of the transformer-based model potential. 
This paper investigates parameter-efficient fine-tuning (PEFT) technology to alleviate overfitting and improve model generalization for 3D models. 
PEFT involves the freezing of the pre-trained backbone that is previously trained on an extensive dataset, while introducing a limited number of learnable parameters to adapt to the new dataset.
This new dataset can be data-rich~\cite{jia2022visual,bahng2022exploring}, few-shot~\cite{LeeDJ2023read}, or long-tailed~\cite{DongB2022lpt}, as PEFT equips the model with knowledgeable priors.
AdapterFormer~\cite{chen2022adaptformer} is an effective PEFT method.
It appends the MLP layer in Eq~(\ref{eq:block}) with a bottleneck module and has been empirically validated for its efficacy in handling 2D image data.
We utilize this architecture for calibrating the point tokens alongside the 2D image attention, namely introducing a trainable bottleneck module.
Formally, the calibration of point embedding is calculated as follows:
\begin{equation}\label{3Dadaptformer}
\begin{cases}
  \tilde{z}_i^{P,(l)} =  \texttt{MSA}^l\left( z_i^{P,(l-1)} \right) + z_i^{P,(l-1)}\\ 
  \hat{z}^{P,(l)}  = \texttt{ReLU}\left( \texttt{LN} (\tilde{z}_i^{P,(l)}) \cdot \mathbf{W}_\text{enc} \right) \cdot \mathbf{W}_\text{dec} \\
  z_i^{P,(l)}  = \texttt{MLP}\left(\texttt{LN} (\tilde{z}_i^{P,(l)}) \right) + s \cdot \hat{z}^{P,(l)}  +  z_i^{P,(l-1)}
\end{cases},
\end{equation}
where $\mathbf{W}_\text{dec} \in \mathbb{R}^{d \times \hat{d}}$ and $\mathbf{W}_\text{dec} \in \mathbb{R}^{\hat{d} \times d}$ are the only learnable parameters for model fine-tuning.
The dimensions satisfy $\hat{d} \ll d$.
All other parameters within the transformer blocks remain fixed.
$s$ is a scale factor.
The input of the first multi-head self-attention block $\texttt{MSA}^1$ is from the sorted point embeddings, namely $ \left[z_i^{P,(0)} \right]  = \mathcal{Z}^{P}_s$. 
The framework of PointFormer is shown in Figure~\ref{fig:PointFormer}.

% \begin{figure}[t]
%     \centering
%     \includegraphics[width=0.65\linewidth, height = 0.85\linewidth]{fig/PointFormer2.pdf}
%     \vspace{6pt}
%     \caption{The framework of PointFormer.}
%     \label{fig:PointFormer}
%     \vspace{18pt}
% \end{figure}

%The preservation of spatial relationships is achieved by:
%1) We partition the input point cloud into point patches. 
%This step endows the point cloud with fundamental separability. 
%As shown in Figure 4(a), after being processed by a random PointNet, points embedding from the same class tend to aggregate.
%2) Point embeddings are obtained using mature techniques, such as PointNet and PointMLP, which have been demonstrated to effectively utilize positional information and preserve spatial properties.

In this way, leveraging spatial relationships by 2D attention mechanisms is achieved by:
1) Since the image tokens fed into the 2D model are arranged in order, we sort the point embeddings according to their corresponding point patch centroids utilizing Morton code.
This process facilitates better attention adaption.
2) PointFormer is utilized to refine attention discrepancies arising from variations in datasets and data structures.

\subsection{Downstream Tasks}

\noindent\textbf{Classification.}
The class token $z_\text{cls}^{P,(l)}$ output by the last block for point embedding can be utilized for classification.
For clarity, we use $z_\text{cls}$ as a shorthand notation for $z_\text{cls}^{P,(l)}$.
The predicted logit of each class is given by the softmax of the final linear layer:
\begin{equation}
    p_i = \dfrac{e^{w_i \cdot z_\text{cls}}}{ \sum_{j=1}^C e^{w_j \cdot z_\text{cls}} },  
\end{equation}
where $w_i$ is the linear classifier weight and $C$ is the total number of classes.
Eventually, the cross-entropy loss can be utilized to calculate the loss function.

\noindent\textbf{Segmentation.}
Segmentation needs to predict a label for each point.
We employ a U-net style architecture, where the APF serves as the point encoder. 
The segmentation head concatenates the output features from transformer blocks within the encoder, succeeded by deconvolution interpolation and multiple MLP layers to facilitate dense prediction. 
Similarly to classification, the softmax cross-entropy is employed as the loss function.

The overall pipeline of APF is shown in Figure~\ref{fig:pipeline}.

\subsection{Comparison with Existing methods.} 

The principal disparity between APF and existing methods lies in that APF demonstrates the viability of adapting 2D priors to 3D feature space with minimal training parameters, rather than relying on specific network architectures.
APF essentially executes \textit{``2D alignment to 3D''}.
%, namely, APF exploits 2D priors by projecting 3D point clouds into single or multiple 2D representations.
The Morton order in the sequencing step aims to mimic the ordered image tokens fed in ViT.
The PointFormer module facilitates the adaption of prior attention in pre-trained ViT.
Most existing methods perform \textit{``3D alignment to 2D''}.
For example, I2P-MAE~\cite{ZhangI2PMAE23} leverages 2D knowledge by projecting 3D point clouds into multiple corresponding 2D images.
P2P~\cite{Ziyi21P2P} transforms the 3D point cloud of an object into a single RGB image.

In addition, APF employ 2D pre-trained models from divergent perspectives compared to existing methods: input and model, respectively. 
For example, P2P focuses on the transformation of 3D point clouds into 2D images at the input level. 
In contrast, APF incorporates PointFormer at the model level, aiming to adapt the self-attention mechanisms (or feature information) embedded within the 2D priors to accommodate
features extracted from point clouds. 

Moreover, existing methods, such as ACT~\cite{dong2023act}, point-MAE~\cite{pangYT22PointMAE}, and I2P-MAE~\cite{ZhangI2PMAE23}, to name a few, necessitate retraining an additional transformer-based network, resulting in additional computational overhead.
Conversely, the training parameters in APF consist of the point embedding and PointFormer modules, which have relatively smaller parameter sizes.

\section{Experiment}
\subsection{Datasets and Basic Settings}
\noindent\textbf{Datasets.}
We perform object classification tasks on ModelNet40~\cite{wuZR20153d} and ScanObjectNN~\cite{uy2019revisiting}. 
For part segmentation, we utilize ShapeNetPart~\cite{yi2016scalable}. 
ModelNet40 is an extensively employed 3D dataset comprising 12,311 CAD models distributed across 40 object categories.
%It is extensively employed in point cloud object classification and surface normal estimation benchmarks. 
We follow the official split with 9,843 objects for training and 2,468 for evaluation for a fair comparison.
ScanObjectNN is a challenging dataset with inherent scan noise and occlusion, which is sampled from the real world with a comprehensive collection of 15,000 scanned objects spanning 15 distinct classes. 
%This dataset poses challenges to classification tasks due to the presence of inherent scan noise and occlusion.
Following previous works, we perform experiments on three variants: OBJ-BG, OBJ-ONLY, and PB-T50-RS.
ShapeNetPart is a meticulously annotated 3D dataset covering 16 shape categories selected from the ShapeNet dataset. 
This dataset is annotated with part-level labels from 50 classes and each category is characterized by 2 to 6 distinct parts.

\noindent\textbf{Implementation Details.}
We follow the settings in \cite{Ziyi21P2P} and \cite{GuoZQLH23JointMAE}, namely the AdamW optimizer in conjunction with the CosineAnnealing scheduler are employed, initializing a learning rate of $5\times 10^{-4}$ incorporating a weight decay of $5 \times 10^{-2}$. 
For point embedding, we explore a lightweight PointNet.
The ViT-Base (ViT-B) version~\cite{Dosovitskiy21vit} is utilized as the pre-trained 2D model in the experiments. 

\begin{table}[tb]
 \caption{Object classification results on ModelNet40.
  *: For P2P, we refer to the results obtained based on ViT-B to ensure a fair comparison.
  $^\dag$: Using a lightweight PointNet for point embedding.}  \label{tab:com_cls_MN40}
  \vspace{6pt}
 \centering  % 显示位置为中间
 \setlength{\tabcolsep}{12pt}
 %\resizebox{1\linewidth}{!}  
 %\renewcommand{\arraystretch}{0.9}
{\begin{tabular} %{m{2cm}| m{1.5cm}<{\centering}|m{1.2cm}<{\centering}}
{l|c|c}
  \toprule
  Methods & Pre-trained modality  & Acc.(\%) \\ 
  \midrule  
  \multicolumn{3}{c}{DNN-based model} \\
  \midrule
  PointNet~\cite{qi2017pointnet} & N/A & 89.2 \\ 
  PointNet-OcCo~\cite{wangHC2021occo} & 3D & 90.1 \\
  PointNet++~\cite{QiNIPS2017pointnet2} & N/A & 90.5  \\ 
  DGCNN~\cite{Wang2019Dynamic} & N/A & 92.9\\ 
  DGCNN-OcCo~\cite{wangHC2021occo} & 3D & 93.0 \\
  %DGCNN-CrossPoint~\cite{AfhamM22CrossPoint} & 2D & 92.1 \\
  KPConv~\cite{ThomasH2019KPConv} & N/A & 92.9 \\ 
  PAConv~\cite{xuM2021paconv}  & N/A &  93.9 \\ 
  PointMLP~\cite{maXQ22PointMLP} & N/A &94.1 \\
  \midrule  
  \multicolumn{3}{c}{Transformer-based model} \\
  \midrule  
  Transformer~\cite{vaswani2017attention}  & N/A & 91.4  \\
  Transformer-OcCo~\cite{wangHC2021occo} & 3D & 92.1 \\
  Point Transformer~\cite{zhao2021point} & N/A & 93.7   \\ 
  PCT~\cite{guoMH2021pct}  & N/A & 93.2 \\
  Point-BERT~\cite{yu2022point}& 3D & 93.2   \\
  Point-MAE~\cite{pangYT22PointMAE} & 3D & 93.8   \\
  P2P$^*$ ~\cite{Ziyi21P2P} & 2D & 92.4   \\ 
  Joint-MAE~\cite{GuoZQLH23JointMAE} & 3D & 94.0 \\
  \midrule
  APF (ours) $^\dag$ & 2D & \textbf{94.2 }\\    
  \bottomrule
  \end{tabular}}
  \vspace{12pt}
\end{table}

\begin{table}[tb]
 \caption{Object classification results on ScanObjectNN.
 ``Trans." abbreviates Transformer.
 *: same with Table~\ref{tab:com_cls_MN40}.
 $\dag$: \textit{w. PointNet} means that using a lightweight PointNet for point embedding.
 $\ddag$: \textit{w. PointMLP} means that using PointMLP for point embedding.}  \label{tab:com_cls_SONN}
 \vspace{6pt}
 \centering  % 显示位置为中间
 \resizebox{1\linewidth}{!}  
{\begin{tabular}{m{2.5cm}| m{1.6cm}<{\centering}|m{1.2cm}<{\centering}|m{1.2cm}<{\centering}|m{1.2cm}<{\centering}}
  \toprule
  Methods & Pre-trained modality  & OBJ-BG & OBJ-ONLY & PB-T50-RS \\ 
  \midrule  
  \multicolumn{5}{c}{DNN-based model} \\
  \midrule
  PointNet~\cite{qi2017pointnet}  & N/A & 73.8 & 79.2 & 68.0 \\  %
  PointNet-OcCo~\cite{wangHC2021occo}  & 3D & - & - & 80.0\\    %
  PointNet++~\cite{QiNIPS2017pointnet2}  & N/A & 82.3 & 84.3 & 77.9\\  %
  DGCNN~\cite{Wang2019Dynamic}  & N/A & 82.8 & 86.2 & 78.1\\      %
  DGCNN-OcCo~\cite{wangHC2021occo}  & 3D & - & - & 83.9 \\     %
  PRA-Net~\cite{ChengSL21PRANet}    & N/A & - & - & 82.1\\       %
  %MVTN~\cite{ChengSL21PRANet}  & N/A & 92.6 & \textbf{92.3} & 82.8 \\    %
  PointMLP & N/A & - & - & 85.2 \\    %~\cite{maXQ22PointMLP}
  \midrule  
  \multicolumn{5}{c}{Transformer-based model} \\
  \midrule  
  Trans.~\cite{vaswani2017attention}   & N/A & 79.9 & 80.6 & 77.2  \\  %
  %Point Transformer~\cite{zhao2021point} & N/A &     \\ 
  %PCT~\cite{guoMH2021pct}  & N/A &   &   &   \\
  Trans.-OcCo~\cite{wangHC2021occo} & 3D & 84.9 & 85.5 & 78.8\\  %
  Point-BERT~\cite{yu2022point} & 3D & 87.4 & 88.1 & 83.1   \\    %
  Point-MAE~\cite{pangYT22PointMAE}  & 3D & 90.0 & 88.3 & 85.2  \\     %
  P2P$^*$~\cite{Ziyi21P2P}   & 2D & - & - & 84.1    \\          %
  Joint-MAE~\cite{GuoZQLH23JointMAE} & 3D & \textbf{90.9} & 88.9 & 86.1 \\      % 
  \midrule
  APF \textit{w. PointNet}$^\dag$  & 2D & 85.5 & 88.4 & 83.1\\    
  APF \textit{w. PointMLP}$^\ddag$ & 2D & 89.9 & \textbf{89.0} & \textbf{87.8}\\   
  \bottomrule
  \end{tabular}}
 %\vspace{-6pt}
\end{table}

\subsection{Comparison Results}

\begin{table*}[tb]
\caption{Few-shot classification results on ModelNet40.} 
\label{tab:com_few_shot} 
\vspace{6pt}
 \centering  % 显示位置为中间
 \resizebox{0.9\linewidth}{!}  
{\begin{tabular}
%{l|c|c|c|c|c}
{m{3cm}| m{1.5cm}<{\centering}| m{2cm}<{\centering}|m{2cm}<{\centering}|m{2cm}<{\centering}|m{2cm}<{\centering}}
  \toprule
  \multirow{2}{*}{Methods} &\multirow{2}{*}[-1ex]{\makecell[c]{Pre-trained \\ modality }}& \multicolumn{2}{c|}{5-way} & \multicolumn{2}{c}{10-way} \\ 
  \cmidrule{3-6}  &   & 10-shot & 20-shot & 10-shot & 20-shot  \\ 
  \midrule  
  \multicolumn{6}{c}{DNN-based model} \\
  \midrule
  PointNet~\cite{qi2017pointnet} & N/A &52.0\ $\pm$\ 3.8 &57.8\ $\pm$\ 4.9 & 46.6\ $\pm$\ 4.3 & 35.2\ $\pm$\ 4.8 \\
  PointNet-OcCo~\cite{wangHC2021occo} & 3D &89.7\ $\pm$\ 1.9 &92.4\ $\pm$\ \textbf{1.6} & 83.9\ $\pm$\ 1.8 & 89.7\ $\pm$\ \textbf{1.5} \\  
  PointNet-CrossPoint~\cite{AfhamM22CrossPoint} & 2D &90.9\ $\pm$\ 4.8 &93.5\ $\pm$\ 4.4 &84.6\ $\pm$\ 4.7 &90.2\ $\pm$\ 2.2 \\
  DGCNN~\cite{Wang2019Dynamic}  & N/A  &31.6\ $\pm$\ 2.8 &40.8\ $\pm$\ 4.6 &19.9\ $\pm$\ 2.1 &16.9\ $\pm$\ 1.5 \\ 
  DGCNN-OcCo~\cite{wangHC2021occo} & 3D  &90.6\ $\pm$\ 2.8 &92.5\ $\pm$\ 1.9 &82.9\ $\pm$\ \textbf{1.3} &86.5\ $\pm$\ 2.2 \\ 
  DGCNN-CrossPoint~\cite{AfhamM22CrossPoint} & 2D  &92.5\ $\pm$\ 3.0 &94.9\ $\pm$\ 2.1 &83.6\ $\pm$\ 5.3 &87.9\ $\pm$\ 4.2 \\
  \midrule  
  \multicolumn{6}{c}{Transformer-based model} \\
  \midrule  
  Transformer~\cite{vaswani2017attention} & N/A &87.8\ $\pm$\ 5.2 &93.3\ $\pm$\ 4.3 &84.6\ $\pm$\ 5.5 &89.4\ $\pm$\ 6.3\\
  Transformer-OcCo~\cite{wangHC2021occo}  & 3D &94.0\ $\pm$\ 3.6 &95.9\ $\pm$\ 2.3 &89.4\ $\pm$\ 5.1 &92.4\ $\pm$\ 4.6\\
  Point-BERT~\cite{yu2022point}   & 3D & 94.6\ $\pm$\ 3.1 &96.3\ $\pm$\ 2.7 &91.0\ $\pm$\ 5.4 &92.7\ $\pm$\ 5.1\\
  Point-MAE~\cite{pangYT22PointMAE}  & 3D &96.3\ $\pm$\ 2.5 & 97.8\ $\pm$\ 1.8 & \textbf{92.6}\ $\pm$\ 4.1 & 95.0\ $\pm$\ 3.0\\
  Joint-MAE~\cite{GuoZQLH23JointMAE}  & 3D &96.7\ $\pm$\ 2.2 & 97.9\ $\pm$\ 1.8 & \textbf{92.6}\ $\pm$\ 3.7 & 95.1\ $\pm$\ 2.6\\
  \midrule
  APF (ours)  & 2D & \textbf{96.9}\ $\pm$\ \textbf{1.8} & \textbf{98.1} $\pm$\ 1.8 & \textbf{92.6}\ $\pm$\ 2.4 & \textbf{95.7} \ $\pm$\ 1.6 \\     
  \bottomrule  \end{tabular}}
  \vspace{12pt}
\end{table*}

\begin{table*}[tb]
\caption{Part segmentation results on ShapeNetPart. $\text{mIoU}_C$ (\%) is the mean of class IoU. $\text{mIoU}_I$ (\%) is the mean of instance IoU. ``Trans." abbreviates for Transformer.}  \label{tab:com_seg}
\vspace{6pt}
\centering  % 显示位置为中间
\resizebox{\linewidth}{!}  
{\begin{tabular}%{l|c c|c c c c c c c c c c c c c c c c}
  {m{2cm} | m{0.6cm}<{\centering} m{0.6cm}<{\centering}| m{0.5cm}<{\centering} m{0.5cm}<{\centering} m{0.5cm}<{\centering} m{0.5cm}<{\centering} m{0.5cm}<{\centering} m{0.5cm}<{\centering} m{0.5cm}<{\centering} m{0.5cm}<{\centering} m{0.5cm}<{\centering} m{0.5cm}<{\centering} m{0.5cm}<{\centering} m{0.5cm}<{\centering} m{0.5cm}<{\centering} m{0.5cm}<{\centering} m{0.5cm}<{\centering} m{0.5cm}<{\centering} }
  \toprule
  Methods & $\text{mIoU}_C$ & $\text{mIoU}_I$ & aero-plane &  bag & cap & car & chair & ear-phone  & guitar & knife & lamp & laptop & motor-bike & mug & pistol & rocket & skate-board & table \\ 
  \midrule  
  \multicolumn{19}{c}{DNN-based model} \\
  \midrule
  PointNet~\cite{qi2017pointnet} & 80.4 & 83.7 & 83.4 & 78.7 & 82.5 & 74.9 & 89.6 & 73.0 & 91.5 & 85.9 & 80.8 & 95.3 & 65.2 & 93.0 & 81.2 & 57.9 & 72.8 & 80.6 \\
  PointNet++~\cite{QiNIPS2017pointnet2} & 81.9 & 85.1 & 82.4 & 79.0 & 87.7 & 77.3 & 90.8 & 71.8 & 91.0 & 85.9 & 83.7 & 95.3 & 71.6 & 94.1 & 81.3 & 58.7 & 76.4 & 82.6 \\ 
  DGCNN~\cite{Wang2019Dynamic} & 82.3 & 85.2 & 84.0 & 83.4 & 86.7 & 77.8 & 90.6 & 74.7 & 91.2 & 87.5 & 82.8 & 95.7 & 66.3 & 94.9 & 81.1 & 63.5 & 74.5 & 82.6 \\
  KPConv~\cite{ThomasH2019KPConv} & 85.1 & 86.4 & 84.6 & 86.3 & 87.2 & 81.1 & 91.1 & 77.8 & 92.6 & 88.4 & 82.7 & 96.2 & 78.1 & 95.8 & 85.4 & 69.0 & 82.0 & 83.6\\
  PAConv~\cite{xuM2021paconv}  & 84.6 & 86.1 & - & - & - & - & - & - & - & - & - & - & - & - & - & - & - & - \\
  PointMLP~\cite{maXQ22PointMLP} & 84.6 & 86.1 & 83.5 & 83.4 & 87.5 & 80.54 & 90.3 & 78.2 & 92.2 & 88.1 & 82.6 & 96.2 & 77.5 & 95.8 & 85.4 & 64.6 & 83.3 & 84.3 \\
  \midrule 
  \multicolumn{19}{c}{Transformer-based model} \\
  \midrule 
  Trans.~\cite{vaswani2017attention} & 83.4 & 85.1 & 82.9 & 85.4 & 87.7 & 78.8 & 90.5 & 80.8 & 91.1 & 87.7 & 85.3 & 95.6 & 73.9 & 94.9 & 83.5 & 61.2&  74.9 & 80.6 \\
  Point Trans.~\cite{zhao2021point} & 83.7 & \textbf{86.6 } & - & - & - & - & - & - & - & - & - & - & - & - & - & - & - \\
  PCT~\cite{guoMH2021pct} & - & 86.4 & 85.0 & 82.4 & 89.0 & 81.2 & 91.9 & 71.5 & 91.3 & 88.1 & 86.3 & 95.8 & 64.6 & 95.8 & 83.6 & 62.2 & 77.6 & 83.7 \\
  Trans.-OcCo~\cite{wangHC2021occo} & 83.4 & 85.1 & 83.3 & 85.2 & 88.3 & 79.9 & 90.7 & 74.1 & 91.9 & 87.6 & 84.7 & 95.4 & 75.5 & 94.4 & 84.1 & 63.1 & 75.7 & 80.8  \\
  Point-BERT~\cite{yu2022point} & 84.1 & 85.6 & 84.3 & 84.8 & 88.0 & 79.8 & 91.0 & 81.7 & 91.6 & 87.9 & 85.2 & 95.6 & 75.6 & 94.7 & 84.3 & 63.4 & 76.3 & 81.5 \\
  %SPoTr~\cite{parkJY2023SPoTr} &85.4 & \textbf{87.2} & - & - & - & - & - & - & - & - & - & - & - & - & - & - & - & - \\
  Point-MAE~\cite{pangYT22PointMAE} & - & 86.1 & 84.3& 85.0 & 88.3 & 80.5 & 91.3 & 78.5 & 92.1 &  87.4 & 96.1 & 96.1 & 75.2 & 94.6 & 84.7 & 63.5 & 77.1 & 82.4 \\
  P2P$^*$~\cite{Ziyi21P2P} & 82.5 & 85.7 & 83.2 & 84.1 & 85.9 & 78.0 & 91.0 & 80.2 & 91.7 & 87.2 & 85.4 & 95.4 & 69.6 & 93.5 & 79.4 & 57.0 & 73.0 & 83.6 \\
  Joint-MAE~\cite{GuoZQLH23JointMAE} & \textbf{85.4} & 86.3 & - & - & - & - & - & - & - & - & - & - & - & - & - & - & - & - \\
  \midrule 
  APF (ours) & 83.4 & 86.1 & 83.6 & 84.8 & 85.4 & 79.8 & 91.3 & 77.0 & 91.4 & 88.4 & 84.4 & 95.5 & 76.3 & 95.3 & 82.5 & 59.5 & 76.1 & 83.5 \\
  \bottomrule
\end{tabular}}
\vspace{12pt}
\end{table*}

\begin{table}[t]
\caption{Impact of each component. The results are obtained on ModelNet40 dataset. RPN: random PointNet, means that the PointNet is frozen with the randomly initialized parameters.}  \label{tab:abla}
\vspace{6pt}
 \centering  % 显示位置为中间
 \resizebox{0.95\linewidth}{!}
 {
  \begin{tabular}{lcc|c}
  %{m{1.5cm} m{1.5cm}<{\centering} m{1.5cm}<{\centering} |m{2.2cm}<{\centering}}
  \toprule 
  Point Embedding & Point Sequencer & PointFormer & Acc. (\%) \\
  \midrule 
  PointNet  &  \ding{55} & \ding{55} & 89.2 (base) \\  
  PointNet  & \ding{51}  & \ding{55} & 93.2 \color{deepred}($\uparrow$ 4.0)\\ 
  PointNet  & \ding{55}  & \ding{51} & 93.5 \color{deepred}($\uparrow$ 4.3)\\
  PointNet  & \ding{51}  & \ding{51}  & \textbf{94.2} \color{deepred}($\uparrow$ 5.0)\\
  RPN    & \ding{51}  & \ding{51}  & 92.2 \color{deepred}($\uparrow$ 3.0)\\
  \bottomrule
 \end{tabular}
 }
%\vspace{-6pt}
\end{table}

\noindent\textbf{Object Classification.}
The results are presented in Tables~\ref{tab:com_cls_MN40} and \ref{tab:com_cls_SONN}.
On ModelNet40, PointNet and Transformer can be seen as the baseline models.
It can be observed that the 3D pre-training OcCo enhances the performance of PointNet and Transformer by 0.9\% and 0.7\%, respectively. 
In contrast, APF exhibits superior performance, outperforming PointNet and Transformer by 5.0\% and 2.8\%, respectively. 
Furthermore, APF surpasses all other counterparts in performance, including the recently proposed Joint-MAE.
On ScanObjectNN, we empirically validate two versions of point embedding methods: PointNet and PointMLP.
PointNet, PointMLP and Transformer are considered as the baseline models in this context. 
APF consistently outperforms the 3D pre-trained model by a large margin.
For example, on the most challenging split, namely PB-T50-RS, the pre-training OcCo improves PointNet by 12.0\%.
In comparison, APF, employing a PointNet embedding, achieves a remarkable gain of 15.1\% over PointNet.
Furthermore, APF, when using PointMLP embedding, exhibits superior performance, surpassing PointMLP by 2.6\% and outperforming other previous arts.
Although APF may not exhibit as robust performance on OBJ-BG compared to the most recently proposed Joint-MAE and Point-MAE, it surpasses the majority of existing methods overall.
For example, on PB-T50-RS, APF with PointMLP outperforms Joint-MAE and Point-MAE by 1.7\% and 2.6\%, respectively.   

\noindent\textbf{Few-shot Classification.}
To demonstrate the generalization capability of the proposed APF, we conduct experiments under few-shot settings, following the common routine~\cite{yu2022point,GuoZQLH23JointMAE}.
The ``$N$-way, $K$-shot" is a conventional configuration, wherein $N$ classes are randomly selected, and each selected class has $K$ training samples and 20 testing samples. 
We repeat each setting 10 times and report the average performance along with the standard deviation.
The results are presented in Table~\ref{tab:com_few_shot}.
In comparison to both 2D and 3D pre-trained models, APF exhibits superior generalization ability in few-shot learning.
For example, APF achieves noteworthy improvements of 2.9\%, 2.2\%, 3.2\%, 3.3\% over Transformer-OcCo in four settings.
Even in comparison to recently proposed SOTA methods, APF consistently shows superior performance.

\noindent\textbf{Part Segmentation.}
Following prior works~\cite{qi2017pointnet,pangYT22PointMAE,GuoZQLH23JointMAE}, we sample 2,048 points from each input instance and adopt the same segmentation head as Point-MAE~\cite{pangYT22PointMAE} and Joint-MAE~\cite{GuoZQLH23JointMAE}.
The results are shown in Table~\ref{tab:com_seg}.
While APF may not have outperformed SOTA methods across both metrics, it achieves commendable overall performance.
In comparison to P2P, which also leverages image priors, APF exhibits superior performance.
Compared to JointMAT, APF exhibits slightly lower performance in terms of $\text{mIoU}_C$ and $\text{mIoU}_I$. 
However, it is worth noting that Joint-MAE requires training from scratch, underscoring the comparatively lower computational overhead of APF.
\subsection{Further Analysis}

\noindent\textbf{Ablation Study.}
We execute a series of controlled experiments to show the impact of each component of APF.
The results are shown in Table~\ref{tab:abla}.
We can observe that each module in APF can improve the baseline method, namely PointNet. 
The integration of Point Sequencer and PointFormer yields the most significant performance enhancement.
It is noteworthy that a random PointNet (RPN) serves merely to align dimensions, lacking the ability to extract meaningful features. 
Nonetheless, APF with RPN still outperforms the vanilla PointNet (92.2\% over 89.2\%), which shows the potential of image prior in 3D domain.

\noindent\textbf{The Impact of 2D Image Prior.}
We design an experiment to evaluate the impact of 2D priors on the 3D point cloud analysis.
We employ ViT-B, pre-trained on ImageNet-21k, to furnish the 2D prior.
A PointNet for input embedding is randomly initialized and subsequently, its parameters are frozen, leaving only the PointFormer in APF as learnable. 
Under this setting (APF \textit{w.} RPN in Table~\ref{tab:analysis}), the object classification accuracies are 92.2\% and 80.1\% on ModelNet40 and ScanObjectNN, respectively.
APF exhibits considerable performance gains over PointNet by only utilizing randomly initialized embedding projection. 
Furthermore, it even outperforms the Transformer trained from scratch while incurring significantly lower training costs compared to training from scratch.
This observation shows that the prior knowledge embedded in the 2D pre-trained model can significantly aid 3D point cloud analysis, even in scenarios where their training sets and data modalities differ. 
Moreover, the PointFormer proves beneficial in calibrating 2D-3D attention.
Involvement of the point cloud projection in training, as indicated by APF with TPN in Table~\ref{tab:analysis}, further improves the performance of APF.
This demonstrates the essential role of a representative point embedding in fully leveraging the potential offered by 2D prior knowledge.

\begin{figure}[!t]
\centering
\subfloat[Unordered Points]{ 
    \includegraphics[width=0.5\linewidth, height=0.4\linewidth]{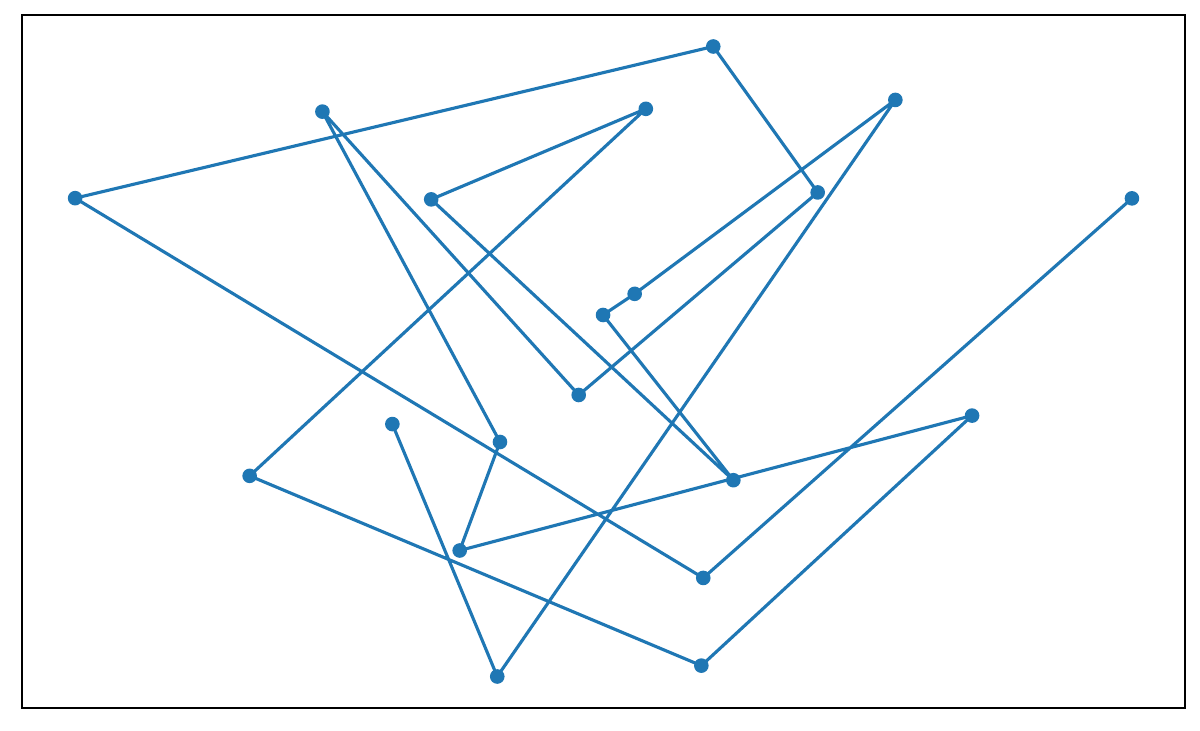}
    \label{fig:unordered}
    }
\subfloat[Ordered Points]{
    \includegraphics[width=0.5\linewidth, height=0.4\linewidth]{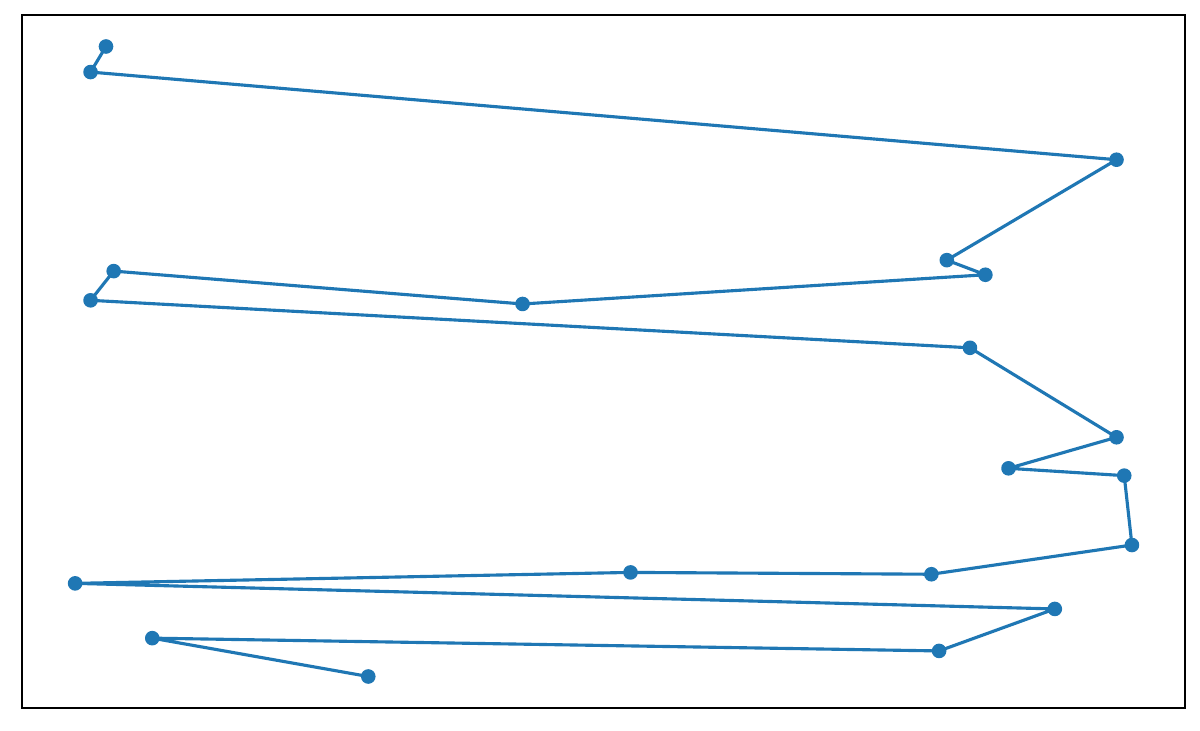}
    \label{fig:ordered}
    }    
    \vspace{6pt}
\caption{Comparison between ordered and unordered. (20 points are selected for clear visualization.)}
\vspace{20pt}
\label{fig:Morton_order}
\end{figure}

\begin{figure}[t]
    \centering
    \includegraphics[width=0.8\linewidth, height=0.6\linewidth]{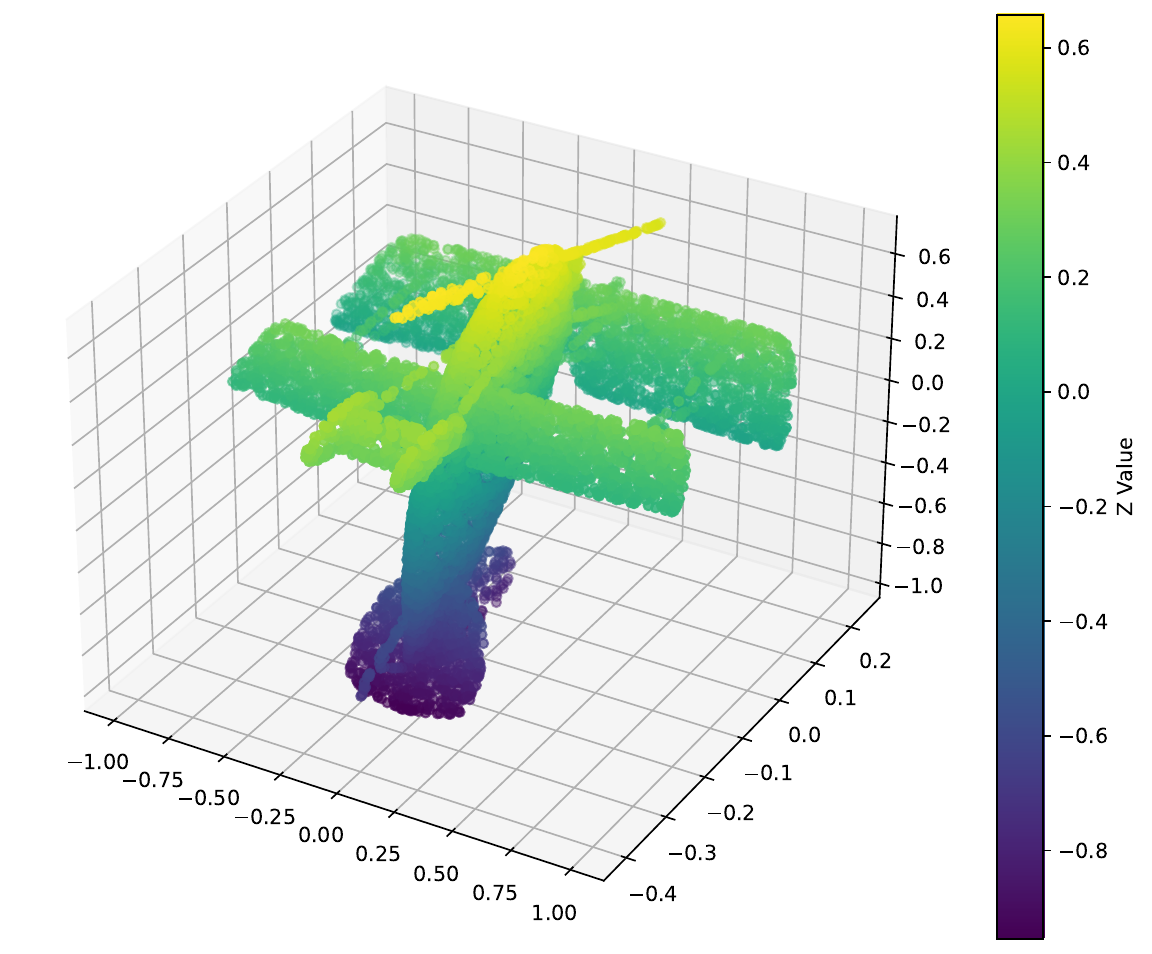}
    \vspace{6pt}
    \caption{Z-order sorting in 3D Space }
    \label{fig:3d_plane}
    \vspace{12pt}
\end{figure}

\noindent\textbf{Technical details for Morton-order.} 
%Point embeddings can be obtained by a lightweight network such as PointNet, and PointMLP.
Sequencing is achieved by Morton code~\cite{morton1966computer}.
In detail, the process begins by selecting one point embedding as the initial point.
Next, the coordinates of the center points corresponding to point embeddings are encoded into one-dimensional space using Morton code, and subsequently sorted to determine their order.
This Morton order, also called Z-order ensures that point embeddings from the closest coordinates are adjacent. 
Figure~\ref{fig:Morton_order} shows an example of the comparison between ordered and disordered point clouds.
We randomly select 20 points for clear visualization.
Figure~\ref{fig:3d_plane} shows the Z-order sorting based on Z-value in 3D space.

\begin{figure}[!tb]
% \begin{figure}
\subfloat[Random PointNet]{ 
    \includegraphics[width=0.5\linewidth, height=0.4\linewidth]{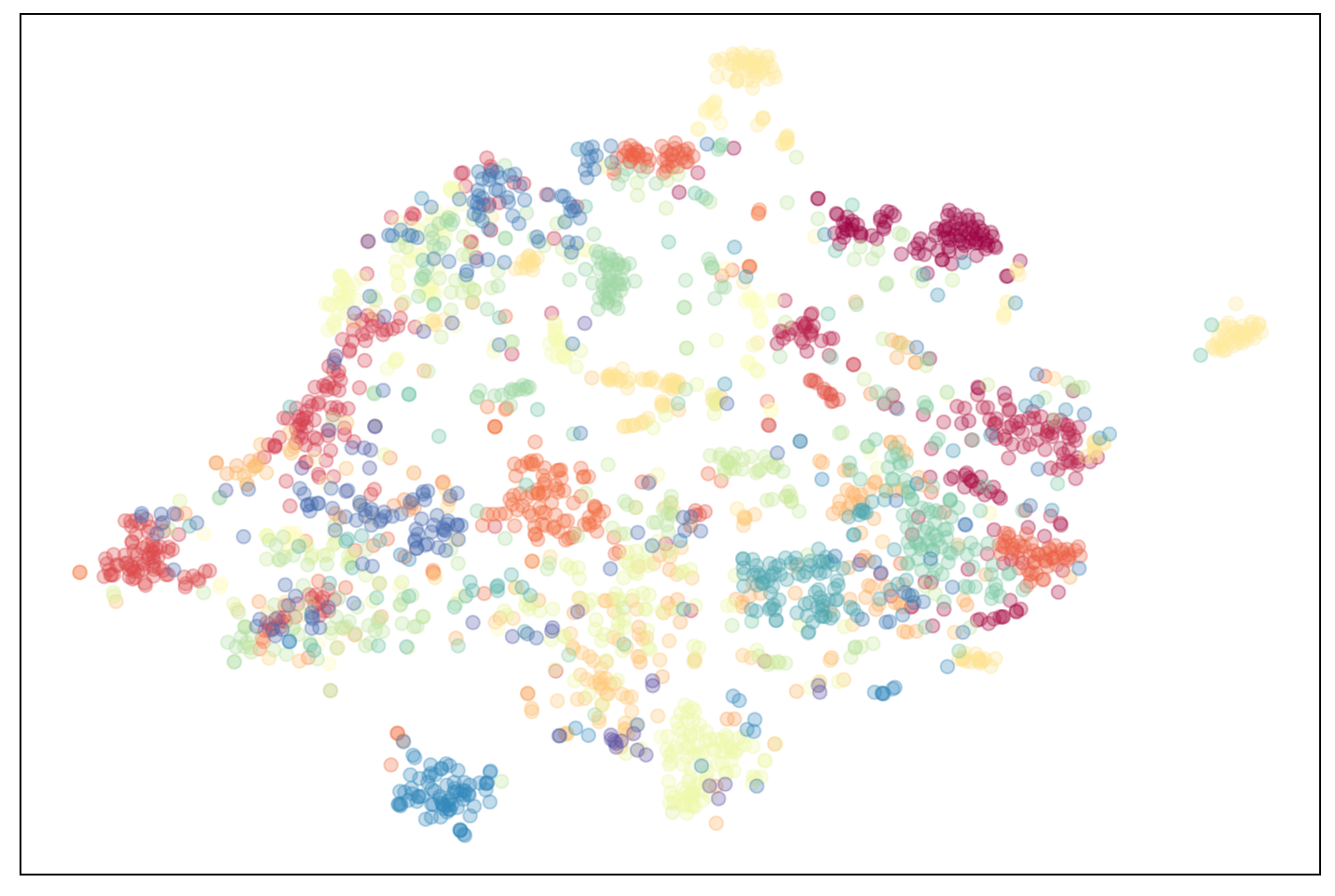}
    \label{fig:RPN}
    }
\subfloat[Random PointNet+PF]{
    \includegraphics[width=0.5\linewidth, height=0.4\linewidth]{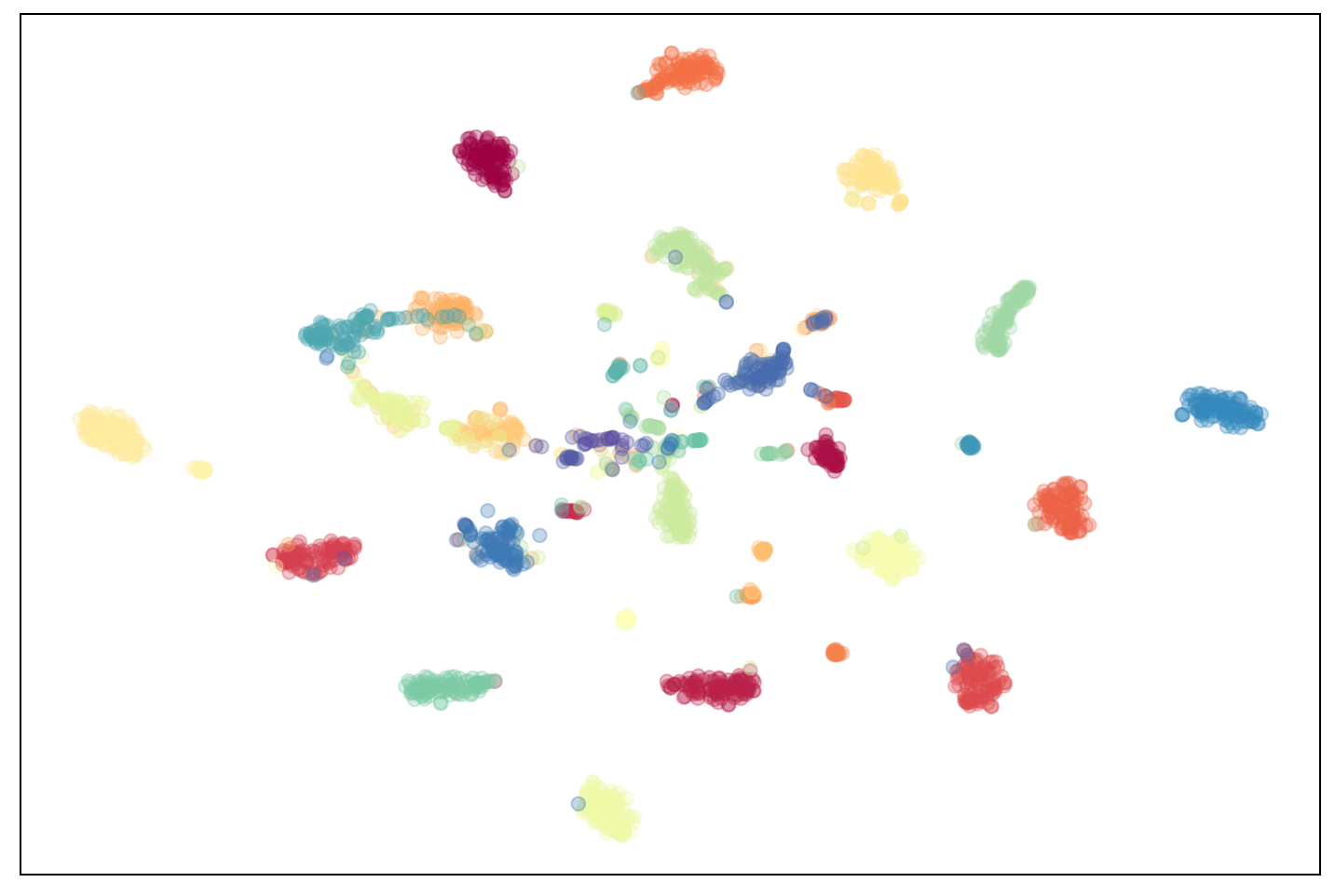}
    \label{fig:FPN_PF}
    }    \\
\subfloat[PointNet]{ 
    \includegraphics[width=0.5\linewidth, height=0.4\linewidth]{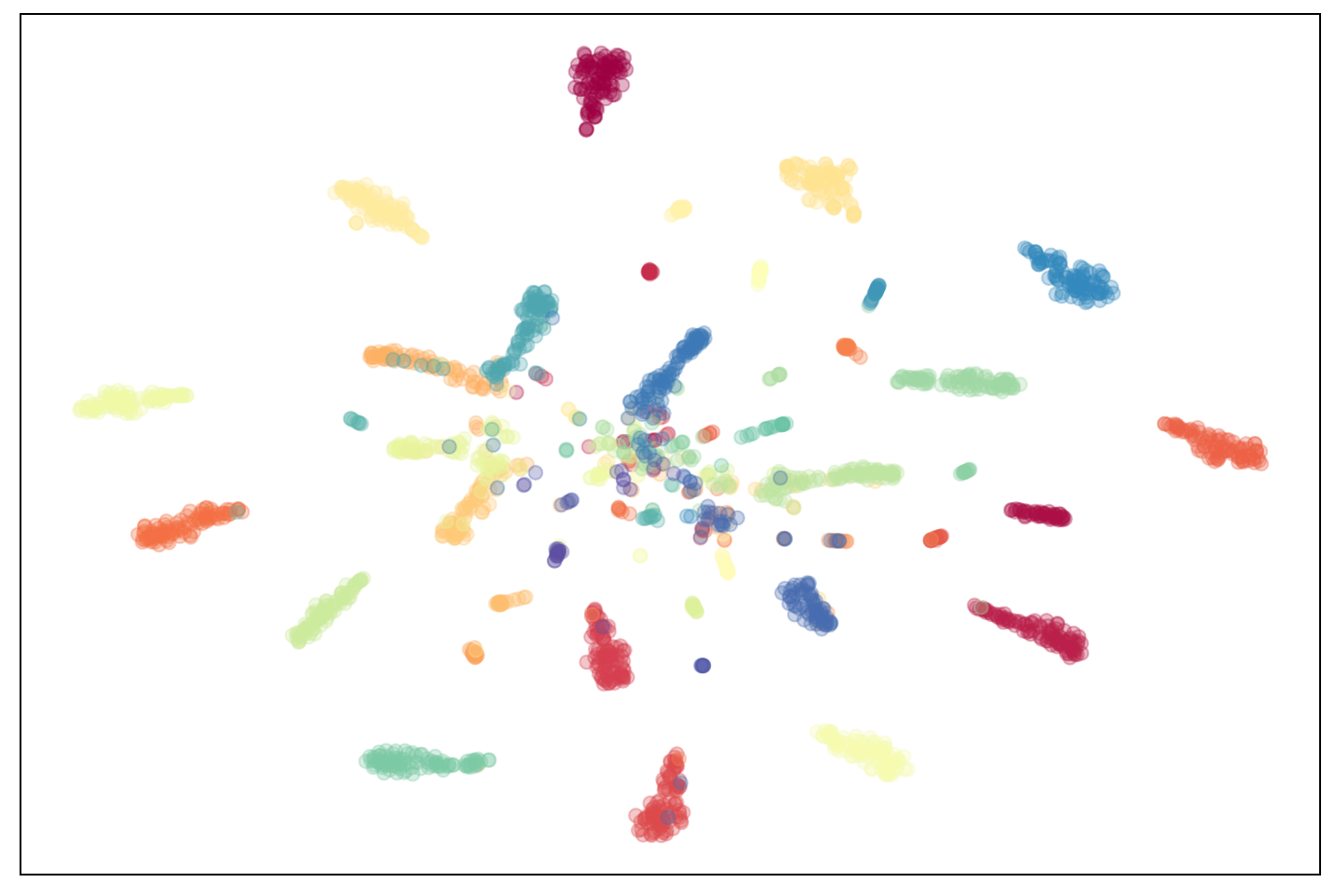}
    \label{fig:PN}
    }
\subfloat[APF]{
    \includegraphics[width=0.5\linewidth, height=0.4\linewidth]{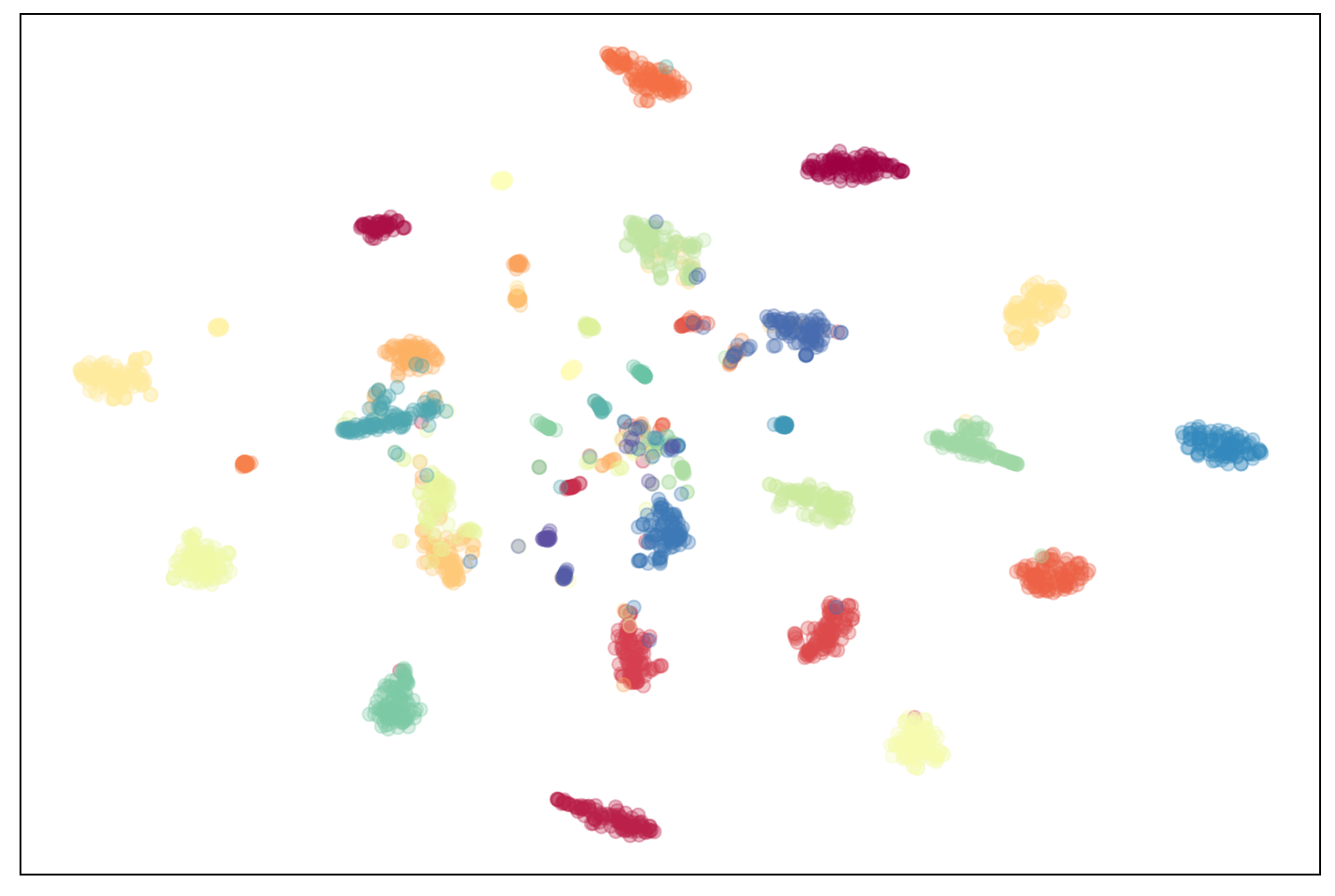}
    \label{fig:ADP}
    }   
\vspace{6pt}
\caption{T-SNE visualization of feature distributions. We show the results on the test set of ModelNet40.}
\vspace{20pt}
% \caption{T-SNE visualization of feature distributions.}
\label{fig:tsne}
\end{figure}

\noindent\textbf{Feature Distributions Visualization.}
We use t-SNE~\cite{van2008visualizing} to visualize feature distributions, which is shown in Figure~\ref{fig:tsne}.
When the dimensions of point clouds are aligned using a random PointNet, features from different classes overlap, as shown in Figure~\ref{fig:RPN}. 
In comparison, the aligned embedding obtained by random PointNet can be evidently separated through PointFormer (PF), as shown in Figure~\ref{fig:FPN_PF}.
This underscores the efficacy of image priors for point cloud analysis.
Similarly, APF further improves the separation of features obtained by the trained PointNet, as shown in Figures~\ref{fig:PN} and \ref{fig:ADP}.

\begin{table}[t]
 \caption{Comparison w.r.t. different training strategies. 
 RPN, short for random PointNet.
 %, namely the PointNet is frozen with the randomly initialized parameters during training.
 TPN, short for trained PointNet, means that the parameters of the lightweight PointNet are also updated during training.
 $\ddag$: PB-T50-RS is utilised.}   \label{tab:analysis}
 \vspace{6pt}
 \centering  % 显示位置为中间
 \resizebox{1.\linewidth}{!}
 {
  \begin{tabular}{ m{2.4cm} m{2.4cm}<{\centering}  m{2.4cm}<{\centering}}
  \toprule 
  Method & Dataset & Acc. (\%) \\
  \midrule 
  PointNet               & ModelNet40  & 89.2   \\
  Transformer            & ModelNet40 & 91.4   \\
  APF \textit{w. RPN} & ModelNet40  & 92.2 \color{deepred}($\uparrow$ 3.0)\\ 
  APF \textit{w. TPN} & ModelNet40  & 94.2 \color{deepred}($\uparrow$ 5.0)\\ 
  \midrule   
  PointNet                & ScanObjectNN$^\ddag$ & 68.0  \\
  Transformer            & ScanObjectNN$^\ddag$ & 77.2   \\
  APF \textit{w. RPN} & ScanObjectNN$^\ddag$  & 80.1 \color{deepred}($\uparrow$ 12.1)\\ 
  APF \textit{w. TPN} & ScanObjectNN$^\ddag$  & 83.1 \color{deepred}($\uparrow$ 15.1)\\ 
  \bottomrule
 \end{tabular}
 }\vspace{9pt}
\end{table}

\noindent\textbf{Quantity of Trainable Parameters.}
\begin{table}[t]
\caption{Comparison with existing methods w.r.t. trainable parameters number. The results are on ModelNet40. ``Pre-tr. Mod." and ``\# Tr. param." are short for ``pre-training modality" and ``parameters number", respectively. $^*$: Reduced-parameter version of APF. }  \label{tab:params}  %$^*$: DGCNN-CP represents DGCNN-CrossPoint.
\vspace{6pt}
 \centering  % 显示位置为中间
 \resizebox{1.\linewidth}{!}
 {
  \begin{tabular}{m{1.8cm} m{1.8cm}<{\centering}  m{1.8cm}<{\centering} m{1.5cm}<{\centering}}    %\raggedleft
  \toprule 
   Method & Pre-tr. Mod.  & \# Tr. param. & Acc. (\%)\\
  \midrule 
  PointNet++  &  N/A & 1.4M & 90.5\\  
  PointMLP & N/A   & 12.6M & 94.1\\ 
   \midrule 
  DGCNN-OcCo & 3D & 1.8M & 93.0\\
  Point-BERT& 3D  & 21.1M & 93.2\\
  Point-MAE & 3D  & 21.1M & 93.8\\
    \midrule 
  %DGCNN-CP$^*$ & 2D  & 1.8M\\
  P2P & 2D  & 0.25M & 92.4\\
  APF (ours) & 2D  & 5.8M & \textbf{94.2}\\   
  APF$^*$ (ours) & 2D  & 2.4M & 93.7\\  
  \bottomrule
 \end{tabular}
 }
\vspace{6pt}
\end{table}
Table~\ref{tab:params} compares the number of trainable parameters with SOTA methods.
In contrast to P2P, our method introduces more parameters during point embedding, yet yields a performance improvement.
In contrast, APF significantly decreases the number of parameters compared to Point-MAE and Point-BERT.
Joint-MAE that achieves SOTA results in part segmentation needs to train a transformer-based network with two branches from scratch, followed by fine-tuning for downstream tasks. 
In contrast, our method requires only direct fine-tuning for downstream tasks, leading to fewer trainable parameters and reduced training costs.
This reduction in parameters, however, is accompanied by a marginal decrease in performance on specific datasets such as ShapeNetPart, which will be the focus of our future research efforts.

\section{Concluding Remarkings}
This paper has initially validated the efficacy of 2D image priors on 3D data using a randomly initialized network for dimension alignment.
The finding demonstrates that pre-trained 2D models can contribute to the analysis of point clouds.
Then, we have proposed the APF framework for fine-tuning the 2D pre-trained visual model on 3D point cloud datasets.
APF consists of a point embedding network for aligning 3D and 2D dimensions, a point sequencer for sorting 3D embedding and the PointFormer for calibrating 2D prior self-attention to 3D embedding space.
APF facilitates the fine-tuning of 2D pre-trained models for 3D point cloud analysis without the need for projecting the 3D point cloud onto a 2D image.

Although APF has demonstrated effectiveness, the performance improvement, in comparison with the existing fine-tuning 2D pre-trained model, is accompanied by a modest increase in the number of training parameters. 
This will be a research focus for improvement in our future work.

\section*{Acknowledgments}
This work was supported in parts by NSFC (62306181, U21B2023, U2001206), Guangdong Basic and Applied Basic Research Foundation (2023B1515120026, 2023A1515110090, 2024A1515010163), DEGP Innovation Team (2022KCXTD025), Shenzhen Science and Technology Program (RCBS20231211090659101), NSFC/RGC (N\_HKBU214/21), RGC GRF (12201321, 12202622, 12201323), RGC SRFS (SRFS2324-2S02).

%%%%%%%%%%%%%%%%%%%%%%%%%%%%%%%%%%%%%%%%%%%%%%%%%%%%%%%%%%%%%%%%%%%%%%%%
%%% Use this command to include your bibliography file.
\clearpage
\bibliography{reference}
\end{document}